\documentclass[10pt,twocolumn,letterpaper]{article}

\usepackage[pagenumbers]{cvpr} % arXiv version

\usepackage{cuted} %< for strip environment

\usepackage{currfile} %< for \currfilebase macro

\usepackage{caption} %< for "\captionof" commands (teaser)
\usepackage{amsmath,amssymb}
\usepackage{booktabs}
\usepackage{array}
\usepackage{graphicx}

\definecolor{cvprblue}{rgb}{0.21,0.49,0.74}
\usepackage[pagebackref,breaklinks,colorlinks,allcolors=cvprblue]{hyperref}

\def\confName{CVPR}
\def\confYear{2027}

\newcommand{\method}{LANTERN}

\title{Can Its Own Misses Teach an MLLM to Find Small Brain Metastases?}

\author{
Yixuan Duan\textsuperscript{1} \quad
Kexin Geng\textsuperscript{1} \quad
Hanwen Xu\textsuperscript{2,\dag} \quad
Wei Qiu\textsuperscript{1,\dag}\\
\textsuperscript{1}Rice University \qquad \textsuperscript{2}University of Pennsylvania\\
{\small \textsuperscript{\dag}Co-last authors}
}

\begin{document}
\maketitle
\begin{abstract}
Multimodal large language models (MLLMs) are becoming general readers of medical images, yet they remain weak at locating small abnormalities.
Brain metastases expose this gap: at 384\,px a 0.1\,mL lesion covers less than a quarter of one visual token, and at one false positive per scan frontier MLLMs find fewer than 30\% of lesions of 0.1--1\,mL.
Fine-tuning on one dataset helps but transfers poorly, and improving it further is slow trial and error steered by a single score.
We introduce \method{} (Lesion-finding AgeNt Through Error-driven Recursive learNing), a framework in which an MLLM improves recursively from its own misses.
\method{} recasts model development as an error-driven loop: a failure census attributes every missed lesion to one of four causes, an LLM mentor revises one of four levels (evidence acquisition, visual representation, learning policy, decision aggregation) per round, and a rule frozen before the first round decides what is kept, without test data and with every tool call audited.
Over 15 unattended rounds, \method{} raises a fine-tuned 4B MLLM from finding 75\% to 94\% of lesions at one false positive per scan on BraTS-MET, and 91\% of those of 0.1--1\,mL, compared with at most 51\% and 29\% for five frontier MLLMs read under the starting protocol; the gains hold on an external cohort (89\%) and on unseen meningiomas (77\%).
These results show that an MLLM's own misses, attributed to their causes, can drive its reliable recursive self-improvement.
\end{abstract}

\section{Introduction}
\label{sec:intro}

% Fig 1 is defined on page 1 so that it floats to page 2; Fig 2 is defined later so that it floats to page 3
\begin{figure*}[t]
\centering
\includegraphics[width=\linewidth]{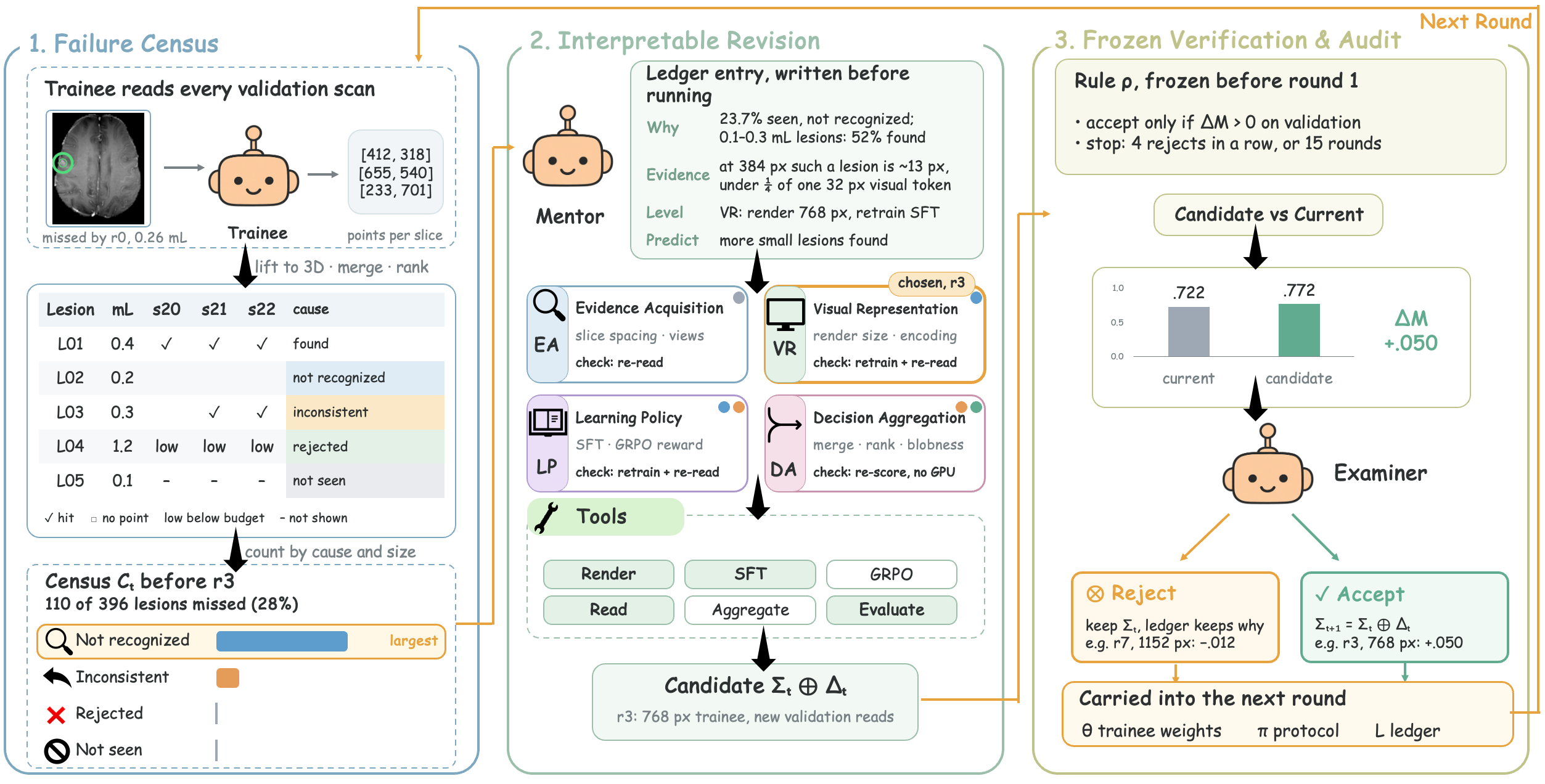}
\caption{\textbf{One round of \method{}}, shown on round 3. (1) The failure census assigns every missed validation lesion one cause. (2) The mentor pre-registers a hypothesis and implements one revision at one of four levels. (3) A frozen examiner accepts or rejects it on validation only (\cref{sec:method}).}
\label{fig:teaser}
\end{figure*}

% ---- Act 1: the problem (MLLM capability) ----
Multimodal large language models (MLLMs) are becoming general readers of medical images: one model, with no task-specific architecture, can describe a radiograph, answer questions about a pathology slide or report on a scan~\cite{li2023llavamed,tu2024medpalmm,moor2023generalist}.
What they still lack is the skill radiology depends on most, saying \emph{where} a small abnormality is.
On medical benchmarks even the strongest MLLMs remain unreliable~\cite{chen2024gmai,hu2024omnimedvqa}, and these benchmarks mostly ask for a label; finding a lesion asks for a location, which is harder.
The usual remedy, training an MLLM on medical or task-specific data, transfers poorly: medically adapted MLLMs do not consistently outperform their general base models~\cite{jeong2024medadapt,hu2024omnimedvqa}, and models trained on data from one institution degrade at others~\cite{zech2018variable}.

Small brain metastases are a demanding test of this skill.
They develop in roughly one in five patients with cancer~\cite{achrol2019brain}, and stereotactic radiosurgery now treats up to ten of them without whole-brain irradiation~\cite{yamamoto2014jlgk0901}, so every lesion has to be found on contrast-enhanced T1-weighted MRI before it can be planned; a lesion that is not found is not treated.
The small ones are the hardest even for dedicated detectors: one built for radiosurgery found 98\% of metastases of at least 6\,mm but 81\% overall~\cite{zhou2020ssd}, and recent systems remain less sensitive below 3\,mm~\cite{topff2025datacentric}.
Frontier MLLMs fall far shorter.
Read slice by slice under the same protocol, Gemini~3.7 Flash reaches $M{=}.46$ on held-out BraTS-MET, Claude Opus~5.5 .41, GPT-6.1 Sol .25, Qwen3.8-27B .25 and Gemma~4 31B .23 (\cref{tab:main}); at 1 FP per scan, the strongest find almost every lesion above 5\,mL but fewer than 30\% of those between 0.1 and 1\,mL.
Medical training does not close this gap: MedGemma 27B, adapted to medical images, scores .002.
Fine-tuning on the target dataset helps but transfers only in part: after supervised fine-tuning on BraTS-MET, a 4B MLLM reaches .747 on its own test set but .641 at another institution.

% ---- Act 2: the gap and the question ----
Teaching an MLLM this skill is usually slow trial and error over coupled choices: what the model is shown, at what resolution, how it is fine-tuned and how its per-slice answers become a per-scan report.
Automation has moved up this stack, from architecture search~\cite{zoph2017nas} and self-configuring pipelines~\cite{isensee2021nnunet} to LLM agents that run machine-learning experiments~\cite{huang2024mlagentbench} and revise their behaviour from feedback~\cite{shinn2023reflexion}, but most are steered by a score.
A score says \emph{whether} a change helped, not \emph{where} the MLLM fails: a lesion lost between surveyed slices, one shown to the model but not recognised and one ranked below the false-positive budget lower $M$ equally, yet each calls for a different fix.
We argue that an MLLM should instead be improved from its own misses, and that doing so recursively faces three obstacles.
\emph{Error attribution}: the usual evaluation reports one number per model, not why each lesion was lost.
\emph{Coupled revisions}: what the MLLM is shown, how it is trained and how its answers are combined interact (thinner slices show more lesions but also yield more false positives; a larger rendering helps only after the model is retrained), so revisions must be free enough to change any of these yet small enough to be attributed.
\emph{Trustworthy inheritance}: every accepted change is built upon by all later rounds, while validation sets are small, retraining alone moves a score by several points~\cite{henderson2018deeprl}, and an agent that sees test data or adjusts its own acceptance rule can produce gains that do not exist~\cite{kapoor2023leakage,varoquaux2022failures}.
We therefore ask: can an MLLM's own misses drive its reliable recursive self-improvement at finding small lesions?

% Table 1 (main results) is defined here so that it floats to the top of page 3
\begin{table*}[t]
\centering\small
\caption{\textbf{Main results.} $M$ = mean recall at 0.25/0.5/1/2 FP per scan; R@1 = recall at 1 FP per scan; Small = recall at 1 FP for lesions of 0.1--1\,mL; lesions $\geq$0.1\,mL. \method{}: mean of 3 seeds. Best in bold, second best underlined.}
\label{tab:main}
\setlength{\tabcolsep}{4pt}
\begin{tabular}{l ccc ccc ccc}
\toprule
& \multicolumn{3}{c}{BraTS-MET test} & \multicolumn{3}{c}{BrainMetShare (external)} & \multicolumn{3}{c}{MEN-RT (unseen type)} \\
\cmidrule(lr){2-4}\cmidrule(lr){5-7}\cmidrule(lr){8-10}
Model & $M$ & R@1 & Small & $M$ & R@1 & Small & $M$ & R@1 & Small \\
\midrule
MedGemma 27B & .002 & .000 & .000 & .003 & .003 & .000 & .007 & .010 & .008 \\
Qwen3-VL-4B zero-shot & .107 & .116 & .004 & .058 & .054 & .016 & .229 & .231 & .000 \\
Gemma~4 31B & .231 & .303 & .104 & .153 & .204 & .103 & .405 & .476 & .040 \\
Qwen3.8-27B & .250 & .323 & .108 & .159 & .213 & .099 & .403 & .455 & .016 \\
GPT-6.1 Sol & .250 & .303 & .065 & .157 & .185 & .063 & .383 & .444 & .040 \\
Claude Opus~5.5 & .409 & .513 & .292 & .290 & .398 & .273 & .480 & .470 & .024 \\
Gemini~3.7 Flash & .460 & .510 & .285 & .342 & .398 & .265 & .521 & .580 & .160 \\
\midrule
\method{} r0 (start) & \underline{.747} & \underline{.747} & \underline{.629} & \underline{.641} & \underline{.641} & \underline{.569} & \underline{.607} & \underline{.646} & \underline{.280} \\
\method{} r15 (ours) & \textbf{.921} & \textbf{.937} & \textbf{.908} & \textbf{.865} & \textbf{.885} & \textbf{.862} & \textbf{.740} & \textbf{.769} & \textbf{.573} \\
\bottomrule
\end{tabular}
\end{table*}

% ---- Act 3: the method ----
We answer with \method{} (\cref{fig:teaser}), with one component per obstacle.
A \emph{failure census} (\cref{sec:census}) attributes every missed validation lesion to exactly one of four causes: \emph{not seen}, \emph{not recognized}, \emph{inconsistent} across training seeds, or \emph{rejected} by the ranking.
A \emph{four-level revision} space (\cref{sec:levels}), namely evidence acquisition, visual representation, learning policy and decision aggregation, lets a mentor, an LLM coding agent, read the census, choose the level itself, pre-register a hypothesis and implement one revision as new code. The mentor wrote the reinforcement-learning trainer, the two-tier aggregation and the lesion-shape score of the final system during the run.
\emph{Frozen verification and audit} (\cref{sec:examiner}) keep a revision only if an examiner, fixed code applying a rule set before round~1, finds that it improves validation performance across three training seeds; the mentor never sees test or external data, and every tool call is audited.
The object being optimised is the MLLM itself, the \emph{trainee} (a 4B Qwen3-VL), together with the protocol by which it reads a scan (\cref{sec:formulation}); every reported lesion is a point the trainee produced, and each accepted system is the starting point of the next round.

% ---- Act 4: results ----
In 15 unattended rounds on BraTS-MET, the examiner accepts 8 revisions spanning all four levels and rejects 7.
On the held-out test set, scored only after the run was frozen, $M$ rises from .747 to .921, above every frontier MLLM we evaluate. At 1 FP per scan, the final system finds 91\% of lesions of 0.1--1\,mL, whereas the strongest frontier MLLMs find fewer than 30\%. Unlike a single round of fine-tuning, the gain carries beyond the training data, to an external cohort (.641 to .865) and to meningiomas, a tumour type the trainee never saw (.607 to .740).
Test misses fall from 100 to 25 of 396 lesions, mostly \emph{not recognized} ones (82 to 18), the cause the mentor targeted most.
Removing any of the seven accepted revisions still in the final system lowers held-out performance, and the mentor arrived, without being told, at choices that mirror radiological practice: thin slices, magnified views and a preference for round, enhancing nodules over tubular vessels.

To summarize, our contributions are (1) a fully unattended recursive self-improvement loop for an MLLM, in which its own misses, sorted by cause, decide each round which level of the system is revised, and an LLM mentor writes each revision as new code; (2) \method{}, which couples a failure census, a four-level revision space and a frozen, audited verification protocol, so that every accepted gain is traceable to one revision and a pre-registered hypothesis and is confirmed on held-out data; and (3) an evaluation over 15 unattended rounds, in which a 4B MLLM improves from $M{=}.747$ to $.921$ on held-out BraTS-MET and finds 91\% of lesions of 0.1--1\,mL at 1 FP per scan, with gains that hold on an external cohort and an unseen tumour type and that surpass frontier MLLMs evaluated under the same protocol.

\section{Related Work}
\label{sec:related}
% Cite only peer-reviewed, formally published papers.

\paragraph{Brain metastasis detection.}
Supervised detectors and segmenters trained on T1c MRI~\cite{liu2017brainmets,charron2018brainmets,bousabarah2020brainmets,xue2020brainmets} reach high sensitivity for large metastases~\cite{grovik2020brainmets}, but it falls for lesions of a few millimetres~\cite{zhou2020ssd,topff2025datacentric}.
Detection is scored by free-response operating characteristics~\cite{chakraborty1990froc}, averaging sensitivity over fixed false-positive rates~\cite{setio2017luna16}; $M$ follows this convention, matching the metric to the clinical question~\cite{maierhein2024metrics}.
These systems are hand-built for one task; we ask how far a general MLLM can be pushed by recursive self-improvement, and compare against MLLMs under a matched protocol.

\paragraph{Multimodal LLMs in medical imaging.}
MLLMs couple a vision encoder with a language model~\cite{alayrac2022flamingo,li2023blip2}, follow visual instructions~\cite{liu2023llava} and may answer with points or boxes~\cite{you2024ferret,deitke2025molmo}. Medical variants add biomedical image--text data~\cite{li2023llavamed,chen2024huatuogptvision,wu2025radfm,tu2024medpalmm} toward generalist medical AI~\cite{moor2023generalist}; promptable segmenters~\cite{kirillov2023sam,ma2024medsam} delineate a structure but do not decide where to look.
Evaluation centres on question answering and classification~\cite{hu2024omnimedvqa,chen2024gmai}, where the strongest models remain unreliable, medical specialisation does not consistently help~\cite{jeong2024medadapt} and models often degrade across institutions~\cite{zech2018variable}. Localising every small lesion in a 3D volume at fixed false-positive budgets, which we score, is rarely evaluated.

\paragraph{Reinforcement fine-tuning and error analysis.}
Reinforcement learning from feedback aligns LLMs~\cite{ouyang2022instructgpt}; group-relative optimisation with verifiable rewards improves reasoning~\cite{guo2025deepseekr1} and transfers to visual grounding~\cite{liu2025visualrft} and medical VQA~\cite{pan2025medvlmr1}. In \method{} it is not the method but one revision the mentor chose (r8).
Decomposing detection errors into types shows researchers where a detector fails~\cite{hoiem2012diagnosing,bolya2020tide}; our census ties causes to the stages of a reading pipeline and makes them an input an agent acts on every round.

\paragraph{Self-improving and automated-research agents.}
Automated machine learning searches architectures~\cite{zoph2017nas} or learning algorithms~\cite{real2020automlzero} or configures pipelines~\cite{isensee2021nnunet}.
LLMs act as optimizers~\cite{yang2024opro}, search programs~\cite{romeraparedes2024funsearch}, write rewards~\cite{ma2024eureka}, accumulate skills~\cite{wang2024voyager}, refine outputs from self-feedback~\cite{madaan2023selfrefine,shinn2023reflexion}, bootstrap their own data or rewards~\cite{zelikman2022star,yuan2024selfrewarding}, run ML experiments~\cite{huang2024mlagentbench,chan2025mlebench}, design agents~\cite{hu2025adas} and recursively improve their own improver~\cite{zelikman2024stop,yin2025godel}; all are steered by a score.
\method{} differs in four ways. \emph{What is improved}: a separate MLLM, the trainee, whose weights and reading protocol the mentor revises, not the agent or its outputs. \emph{What steers a revision}: a census that assigns every missed lesion one cause tied to the level that can remove it, not a score. \emph{What a revision may be}: new code at one of four levels, not a point in a predefined space; the RL trainer, two-tier aggregation and blobness score of the final system did not exist at the start. \emph{What decides acceptance}: a rule fixed before round 1 and applied by code over three seeds, with test and external data absent from the mentor's environment and every tool call audited.

\section{Methods}
\label{sec:method}

\method{} improves an MLLM at finding lesions by repeating one round: it explains every lesion the current MLLM-based system misses, proposes one revision aimed at those misses, and keeps the revision only if a fixed examiner finds that it helps. We first define the task, the metric and the round (\cref{sec:formulation}), then the starting system (\cref{sec:r0}), and then the three parts of a round: the failure census (\cref{sec:census}), the four-level revision (\cref{sec:levels}) and frozen verification (\cref{sec:examiner}). \Cref{sec:r15} describes the system the run ended with.

\paragraph{One round, by example.} \Cref{fig:teaser} shows round 3. \emph{(1) Census:} the trainee reads every validation scan, its points are lifted to 3D, merged and ranked, and every lesion missed at 1 FP per scan is assigned one cause by comparing seed 20 with its sibling seeds; before r3, 110 of 396 validation lesions were missed, 94 of them seen but not recognized, mostly small. \emph{(2) Revision:} the mentor writes a ledger entry (cause, evidence, level, predicted effect) before running anything, picks one level and implements the revision with its tools; in r3 it rendered views at 768 instead of 384\,px and retrained the trainee, because a 0.1--0.3\,mL lesion covered less than a quarter of one visual token. \emph{(3) Verification:} the examiner, fixed code rather than an LLM, compares the candidate with the current system on validation only, under a rule frozen before round 1. An accepted revision ($\Delta M{=}{+}.050$ in r3) is carried into the next round; a rejected one (\eg r7, 1152\,px, $-.012$) is discarded and only its ledger entry is kept.

\begin{figure*}[t]
\centering
\includegraphics[width=0.86\linewidth]{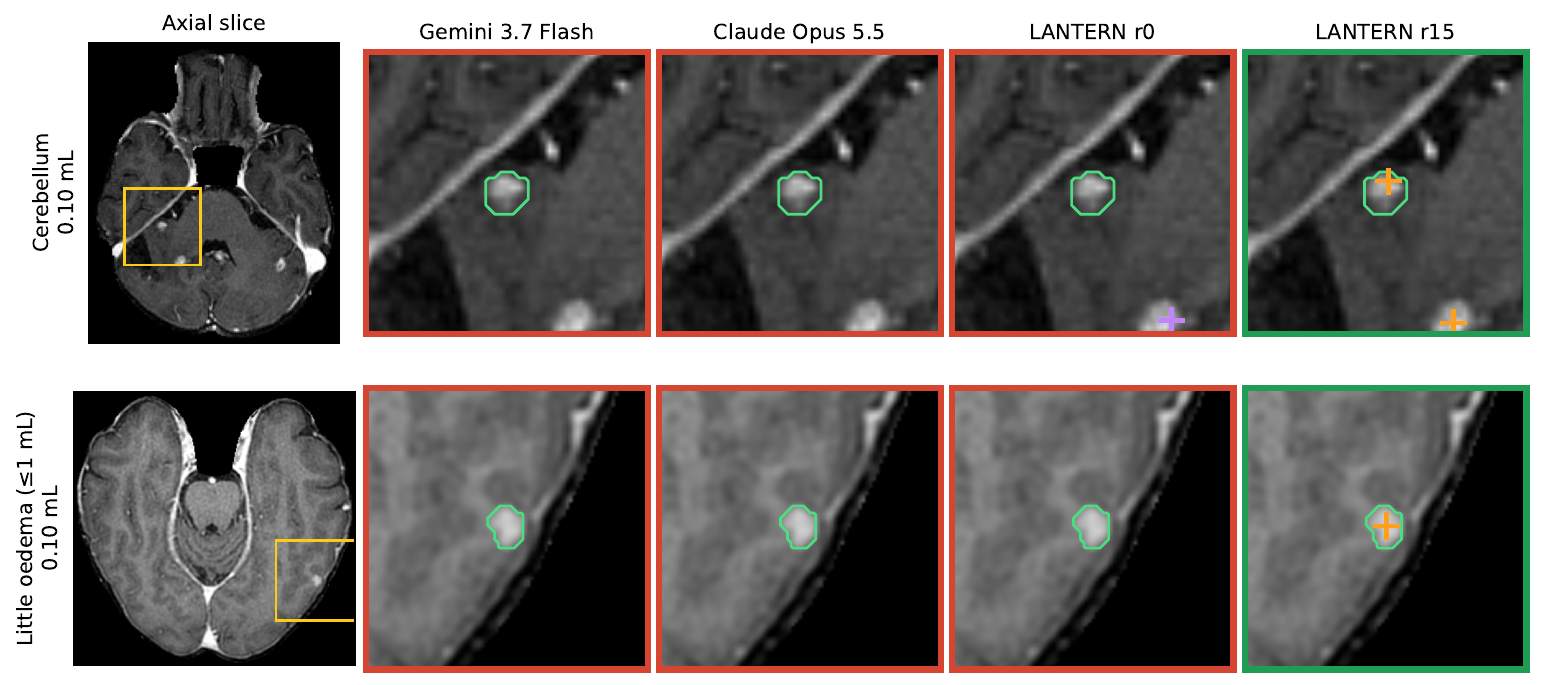}
\caption{\textbf{Hard lesions found only by r15} (test set, T1c). For each group, the smallest lesion that r15 finds at 1 FP per scan while r0, Gemini~3.7 Flash and Claude Opus~5.5 all miss it (seed 20 for \method{}). Left: axial slice through the lesion centroid; yellow box: the 40\,mm field of the zooms. Zooms: green outline, reference lesion; $+$, candidates each system keeps at 1 FP per scan within two slices of the centroid; frame green, found; red, missed.}
\label{fig:hard}
\end{figure*}

\subsection{Problem Formulation}
\label{sec:formulation}

\paragraph{Task.} Given a contrast-enhanced T1-weighted (T1c) brain MRI volume, the system reports a ranked list of 3D points, one per suspected metastasis. Lesions are the 3D connected components (26-connectivity) of the enhancing and non-enhancing tumour labels of the reference segmentation; peritumoral oedema is excluded. A lesion counts if its volume is at least 0.1\,mL. A reported point is a \emph{hit} if it lies within 5\,mm of the lesion mask, so a point is credited for finding a lesion, not for delineating it. Points within 5\,mm of a lesion smaller than 0.1\,mL are ignored, neither hits nor false positives; every other point that is not a hit is a false positive (FP).

\paragraph{Metric.} Sweeping a threshold over the ranked points of all scans gives a free-response operating curve. Let $\mathrm{Rec}(b)$ be the highest lesion-level recall reached with at most $b$ FPs per scan. Following the competition performance metric for lung nodules~\cite{setio2017luna16}, we summarise the curve by the mean recall at four budgets:
\begin{equation}
M=\tfrac14\textstyle\sum_{b\in\{0.25,\,0.5,\,1,\,2\}}\mathrm{Rec}(b).
\label{eq:metric}
\end{equation}

\paragraph{Optimisation target: an MLLM.} The object \method{} optimises is an MLLM $f_\theta$ with weights $\theta$, which maps one image and a text prompt to a text answer, here a list of 2D points. Because $f_\theta$ reads 2D images, it needs a reading protocol $\pi$ to report on a 3D volume: $\pi$ decides which views of the volume $f_\theta$ is shown and how they are rendered, how many answers are drawn per view, and how the answers of one scan are lifted to 3D, merged and ranked. The detector is therefore the MLLM read under its protocol, $\pi\circ f_\theta$, and the goal is
\begin{equation}
\max_{\theta,\,\pi}\; M\big(\pi\circ f_\theta;\,\mathcal{D}_\text{test}\big)
\quad\text{using only } \mathcal{D}_\text{train} \text{ and } \mathcal{D}_\text{val},
\label{eq:objective}
\end{equation}
where $\mathcal{D}_\text{train}$, $\mathcal{D}_\text{val}$ and $\mathcal{D}_\text{test}$ are the training, validation and test sets. Two constraints keep the MLLM at the centre. First, the model class of $f$ is fixed: $\theta$ may be changed by any training procedure, but the MLLM is never replaced by, or combined with, a task-specific detection network. Second, every point in the final report is a point that $f_\theta$ produced; $\pi$ may choose, merge and rank the MLLM's answers, but never creates a lesion candidate of its own. A gain in $M$ therefore comes from the MLLM’s own reports, as chosen, merged and ranked by $\pi$.

\paragraph{System state and one round.} \method{} optimises $\theta$ and $\pi$ jointly, not only the weights of the MLLM. The state before round $t$ is $\Sigma_t=(\theta_t,\pi_t,\mathcal{L}_t)$: the trainee's weights $\theta_t$, the protocol $\pi_t$ that turns a scan into a ranked list (which views are read, how they are rendered, how reads are aggregated), and the ledger $\mathcal{L}_t$, the written record of every past hypothesis and its outcome. One round is
\begin{align}
C_t &= \textsc{Census}(\Sigma_t;\,\mathcal{D}_\text{val}), \label{eq:census}\\
\Delta_t &= \textsc{Mentor}(\Sigma_t, C_t), \quad \ell(\Delta_t)\in\{\text{EA},\text{VR},\text{LP},\text{DA}\}, \label{eq:mentor}\\
v_t &= \textsc{Examine}(\Sigma_t,\Delta_t;\,\rho), \label{eq:examine}\\
\Sigma_{t+1} &= \begin{cases} (\Sigma_t\oplus\Delta_t)\cup\{(\Delta_t,v_t)\} & v_t=\text{accept},\\ \Sigma_t\cup\{(\Delta_t,v_t)\} & v_t=\text{reject},\end{cases} \label{eq:update}
\end{align}
where $\mathcal{D}_\text{val}$ is the validation set, $\ell(\Delta_t)$ the level of the revision, $\rho$ the acceptance rule, $\oplus$ applies the revision to $(\theta,\pi)$, and $\cup\{\cdot\}$ appends the hypothesis and verdict to the ledger. The rule $\rho$, the evaluation code and the data splits are fixed before round 1 and cannot be changed by the loop. Each round therefore starts from the last accepted system and from everything learned so far, including what was learned from rejected revisions; this is the recursion. The run stops after 15 rounds or after 4 consecutive rejections.

\paragraph{Roles.} The \emph{trainee} is the MLLM $f_\theta$ being optimised (\cref{sec:r0}); the \emph{mentor} is an LLM coding agent (Claude Opus~5.5, weights unchanged) that reads the census, writes the hypothesis and implements the revision; the \emph{examiner} is fixed code that applies $\rho$ (\cref{sec:examiner}). No human chooses, edits or rescues a revision.

\subsection{Trainee and Starting System}
\label{sec:r0}

\paragraph{Views.} The trainee reads one axial slice at a time. Each volume is normalised by clipping intensities to the 0.5th--99.5th percentile of non-zero voxels and rescaling to $[0,1]$. A slice is cropped to the brain bounding box plus a 5\,mm margin, shown in radiological orientation and resized so that its long side is $L$ pixels. A \emph{survey} reads slices at a fixed spacing $s$ (in mm) through the brain.

\paragraph{Reading.} The trainee is Qwen3-VL-4B-Instruct. For each view it receives the image and a fixed prompt that names the sequence and the slice position ($z$ in mm and its index in the survey) and asks for the centre of every visible metastasis as JSON points in 0--1000 normalised image coordinates, or an empty list. Each reported point is mapped back to a voxel through the crop and the slice index, so the reads of one scan become a set of 3D points.

\paragraph{Training.} r0 is trained by supervised fine-tuning (SFT) on reports derived from the training masks. Each 5\,mm survey slice of a training scan, plus one copy shifted by 1--4 slices, becomes one example whose target lists the in-slice centroid of every lesion cross-section of at least 3 voxels (lesions $\geq$0.1\,mL), and is empty otherwise. We train LoRA adapters~\cite{hu2022lora} on the language model with the vision encoder frozen (hyperparameters in Supp.~Sec.~\ref{sec:supp_impl}).

\paragraph{Aggregation.} Points of one scan are linked when they are within a merge radius $r$ of each other, and each connected group becomes a candidate located at its medoid, a point that was actually reported, not an average. A candidate's \emph{support} is the number of distinct slices that reported it, and candidates are ranked by support.

\paragraph{r0.} The starting system reads a 5\,mm survey at $L{=}384$ with one greedy read per view and aggregates with $r{=}12$\,mm. We train three trainees with different seeds (20, 21, 22); every number we report for \method{} is either per seed or their mean.

\subsection{Failure Census}
\label{sec:census}

The census turns ``the system missed some lesions'' into a count of \emph{why} it missed them. It is computed on the validation set at a budget of 1 FP per scan. For the system of seed $s$ and a lesion $j$ that it misses at this budget, we test four conditions in order and assign the first that holds:
\begin{enumerate}
  \item \emph{Not seen}: no surveyed slice intersects lesion $j$. The trainee was never shown it.
  \item \emph{Rejected}: a candidate of seed $s$ matches $j$, but ranks below the threshold that keeps 1 FP per scan. The trainee found the lesion; aggregation discarded it.
  \item \emph{Inconsistent}: no candidate of seed $s$ matches $j$, but a candidate of a sibling seed $s'\neq s$, trained and read in the same way, does. The lesion is recognisable by this recipe, but not reliably.
  \item \emph{Not recognized}: otherwise. The lesion was shown, and no trainee of this recipe reported it.
\end{enumerate}
Every missed lesion receives exactly one cause, so the four counts add up to the misses at the budget. The census reports each count overall and by lesion size (0.1--0.3, 0.3--1, 1--5 and $>$5\,mL), with diagnostics listed in Supp.~Sec.~\ref{sec:supp_impl}. Supp.~\cref{fig:failure} shows a real validation example of each cause. The mentor may write new diagnostics; it did so seven times in the run.

\subsection{Four-Level Revision}
\label{sec:levels}

A revision may change anything the trainee or its protocol does, but it is assigned to exactly one of four levels. Each level is the natural remedy for one or two failure causes (Supp.~\cref{tab:levels}), so the census points the mentor to where a revision is likely to help without prescribing what the revision is.

\paragraph{Evidence Acquisition (EA)} decides what the trainee is shown: slice spacing, sequences, planes and the number of reads per view; more views also give more chances to recognize a lesion. \paragraph{Visual Representation (VR)} decides which image features the trainee can extract from a view: rendering resolution and how the image is encoded, for example with neighbouring slices as colour channels. A new input usually requires retraining, which belongs to the same revision. \paragraph{Learning Policy (LP)} decides how the trainee is trained: training data, loss and reward. \paragraph{Decision Aggregation (DA)} decides how reads become one ranked report: merge radius, candidate scores and how several reads of a view are combined. A DA revision can be tested on saved reads and needs no GPU.

\paragraph{The mentor's protocol.} In each round the mentor runs the census on the current accepted system and chooses one revision at one level, aimed at the target it judges most valuable. \emph{Before submitting any job}, it appends to the ledger the failure explanation, the hypothesis, the level, the predicted effect and how the revision will be judged. It then writes the code, submits the training or reading jobs, calls the examiner and appends the outcome. The mentor starts with tools to render, read, build SFT data, train with SFT, aggregate, evaluate and compare systems, and may change or add any of them except the evaluation core (\cref{sec:examiner}).

\subsection{Frozen Verification and Audit}
\label{sec:examiner}

\paragraph{Acceptance rule.} Let $M_s(\Sigma)$ be the validation $M$ of system $\Sigma$ with the trainee of seed $s$, and $\Delta M_s=M_s(\Sigma_t\oplus\Delta_t)-M_s(\Sigma_t)$, which compares each proposed seed with the accepted system of the same seed. The examiner applies
\begin{equation}
v_t=\text{accept}\iff
\begin{cases}
\Delta M_{20}>0 & \text{DA on saved reads},\\
\overline{\Delta M}>0 & \text{otherwise},
\end{cases}
\label{eq:rule}
\end{equation}
where $\overline{\Delta M}$ is the mean of $\Delta M_s$ over seeds 20, 21 and 22.
Revisions that retrain the trainee or read new views are judged on all three seeds, since retraining alone can move a single seed by several points. A DA revision only re-ranks reads that already exist, so its effect is deterministic given the reads, and it is judged on the first seed; the other seeds are scored and logged but do not decide. Every revision tried counts as a round, a rejected result may not be re-run with new seeds, and a job may be repeated only after an infrastructure failure, which must be logged.

\paragraph{What the mentor can and cannot touch.} The mentor works in its own copy of the project that contains only the training and validation sets; the test set and the external cohorts are not present. It has no GPU and no direct access to the cluster: it writes job files that a gateway submits (at most 24\,h, 128\,GB and one GPU per job). The evaluation core (point-to-lesion matching, FP counting, the operating curve and \cref{eq:metric}) is hashed when the run starts. The mentor may write new evaluation tools but may not change the core.

\paragraph{Human role and audit.} Humans fix the rule, the splits and the evaluation before round 1 and do not change them afterwards. During the run, humans may only repair infrastructure faults (\eg a job submitted twice by the gateway), and each repair is written to a separate infrastructure log. After the run, an audit of all 373 tool calls of the mentor confirms that the evaluation core is unchanged, that every job read only training or validation data, and that the new aggregation scores use the T1c image but no labels. Test and external data are scored once, after the run is frozen, and never inform a decision.

\subsection{The System the Run Ended With}
\label{sec:r15}

The run stopped at the 15-round limit with r15, which differs from r0 at every level. \Cref{tab:ablation} lists the round that introduced each component.

\paragraph{EA and VR.} r15 reads a 2\,mm survey instead of 5\,mm, and renders views at $L{=}768$ instead of 384, with the SFT data rebuilt and the trainee retrained at that size.

\paragraph{LP.} After SFT, the trainee is trained for 600 steps with group-relative policy optimisation~\cite{guo2025deepseekr1}. Each step draws 16 training views (half with a lesion), samples $G{=}4$ reports per view at temperature 1, and rewards report $i$ with
\begin{equation}
R_i=N_\text{hit}-\lambda\,N_\text{FP},
\label{eq:reward}
\end{equation}
where $N_\text{hit}$ is the number of lesions $\geq$0.1\,mL hit and $N_\text{FP}$ the number of points more than 5\,mm from every lesion, with $\lambda{=}1$ and the same 3D hit definition as the evaluation; the advantage is $R_i$ minus its group mean (optimisation details in Supp.~Sec.~\ref{sec:supp_impl}).

\paragraph{DA.} Each view is read once greedily and 4 times by sampling at temperature 1. Candidates from the greedy read are merged at $r{=}6$\,mm and scored by
\begin{equation}
\mathrm{score}(c)=\mathrm{support}(c)+0.9\,\frac{\beta(c)}{\beta(c)+0.05},
\label{eq:score}
\end{equation}
where $\beta(c)$ is the largest 3D Hessian blobness~\cite{frangi1998multiscale,lindeberg1998scale} of the T1c volume within 3\,mm of the candidate, so a round, bright nodule ranks above a tubular vessel with the same support (definition in Supp.~Sec.~\ref{sec:supp_impl}). Candidates found only by the sampled reads, more than 6\,mm from every greedy candidate, are added on the same scale, each (slice, sample) report counting $1/G{=}0.25$ of a greedy slice report.

\section{Experiments and Results}
\label{sec:results}

\begin{table}[t]
\centering\small
\caption{\textbf{Finer visual measures}, test set. J: slice-level Youden index. Dist.: median hit-to-centroid distance (mm). D50: diameter (mm) found with 50\% probability. Recall at 1 FP per scan in scans with $\geq$10 lesions, near the surface (Surf., $\leq$10\,mm), deeper (Deep) and in the lowest-contrast quartile (Low c.). Open-weight MLLMs in Supp.~Tab.~\ref{tab:vision_full}.}
\label{tab:vision}
\setlength{\tabcolsep}{1.8pt}
\resizebox{\linewidth}{!}{%
\begin{tabular}{@{}lcccccccc@{}}
\toprule
 & J$\uparrow$ & \multicolumn{2}{c}{Dist.$\downarrow$} & D50$\downarrow$ & $\geq$10$\uparrow$ & Surf.$\uparrow$ & Deep$\uparrow$ & Low c.$\uparrow$ \\
\cmidrule(lr){3-4}
Model & & all & $<$1\,mL & & & & & \\
\midrule
GPT-6.1 Sol & .239 & \underline{1.92} & \underline{1.54} & 7.7 & .212 & .221 & .348 & .293 \\
Claude Opus~5.5 & .552 & 2.69 & 2.88 & 9.7 & .325 & .393 & .578 & .475 \\
Gemini~3.7 Flash & .687 & 2.88 & 2.99 & 9.7 & .300 & .443 & .547 & .465 \\
\midrule
\method{} r0 & \underline{.706} & 2.02 & 1.74 & \underline{7.1} & \underline{.633} & \underline{.624} & \underline{.814} & \underline{.609} \\
\method{} r15 & \textbf{.798} & \textbf{1.22} & \textbf{0.99} & \textbf{4.9} & \textbf{.908} & \textbf{.893} & \textbf{.961} & \textbf{.875} \\
\bottomrule
\end{tabular}}
\end{table}

\subsection{Experimental Setup}
\label{sec:setup}

\paragraph{Datasets.} \emph{BraTS-MET 2024}~\cite{moawad2025bratsmets}, part of the BraTS benchmark series~\cite{menze2015brats}, provides T1c scans with expert segmentations. We split it by patient into 60\% training, 20\% validation and 20\% test, giving 393, 130 and 129 studies; the validation and test sets each contain 396 lesions $\geq$0.1\,mL. Only the training and validation sets are visible to the mentor. \emph{BrainMetShare}~\cite{grovik2020brainmets} is an external cohort from a different institution and scanner, with 105 studies and 314 lesions $\geq$0.1\,mL; an image fingerprint check finds no overlap with BraTS-MET. \emph{BraTS-MEN-RT 2024}~\cite{labella2026bratsmenrt} contains meningiomas, a tumour type the trainee was never trained on, in 500 studies with 615 lesions $\geq$0.1\,mL; scans are skull-stripped with SynthStrip~\cite{hoopes2022synthstrip}. Both external sets are read zero-shot, without retraining.

\paragraph{Baselines.} We compare with multimodal LLMs, since the question is how far a model of this kind can be improved: zero-shot Qwen3-VL-4B (the trainee's own backbone), Qwen3.8-27B, Gemma~4 31B, MedGemma 27B, Claude Opus~5.5, GPT-6.1 Sol and Gemini~3.7 Flash. Every baseline reads exactly like r0, with the same 5\,mm survey, 384\,px views, prompt and one read per view, and is aggregated and scored by the same code.

\paragraph{Evaluation.} We report $M$ (\cref{eq:metric}), recall at each budget, and recall at 1 FP per scan by lesion size. \method{} is reported per seed and as the mean over the three seeds.

\begin{figure}[t]
\centering
\includegraphics[width=0.85\linewidth]{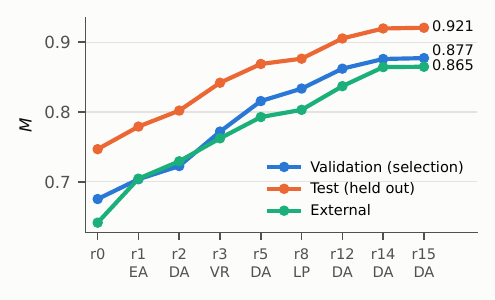}
\caption{\textbf{$M$ of every accepted system} (mean of 3 seeds). Only validation was used for selection.}
\label{fig:traj}
\end{figure}

\subsection{Comparison with Frontier MLLMs}
\label{sec:exp}

\paragraph{Recursive self-improvement lets a 4B MLLM far outperform frontier MLLMs.} On the held-out test set, r15 reaches $M{=}.921$ (seeds .929/.920/.914), compared with .747 for r0, an improvement of about .17 on every seed (\cref{tab:main}). The strongest frontier MLLM, Gemini~3.7 Flash, reaches .460: it finds 88\% of lesions when false positives are unlimited, but needs 4.2 false positives per scan to do so. Claude Opus~5.5 reaches .409 when it reads the scans itself, whereas the same LLM, acting as the mentor, raised the 4B trainee to .921. The medical MLLM MedGemma 27B scores near zero (.002), with 61 false positives per scan.

\paragraph{The gain holds at another institution and on an unseen lesion type.} On BrainMetShare, r15 reaches .865 compared with .641 for r0, a gain of .19 to .25 on each seed. On BraTS-MEN-RT meningiomas, read without retraining, r15 reaches .740 (seeds .750/.739/.732) compared with .607 for r0 and at most .521 for a frontier MLLM (Gemini~3.7 Flash).

\paragraph{The gain is largest on the smallest lesions.} At 1 FP per scan, r15 finds 90.8\% of lesions of 0.1--1\,mL, compared with 62.9\% for r0 and 29.2\% for Claude Opus~5.5. On the 115 lesions of 0.05--0.1\,mL, which lie below the main threshold and never served as training targets, recall rises from .151 to .638 (all sizes and datasets in Supp.~Tab.~\ref{tab:size}).

\paragraph{r15 also sees more precisely.} r15 leads r0 and every frontier MLLM on each finer visual measure in \cref{tab:vision}. Its hits lie a median 1.2\,mm from the lesion centroid, it detects half of all lesions 4.9\,mm across (7.1\,mm for r0), and it keeps .91 recall in scans with ten or more lesions, where r0 falls to .63 and the best frontier MLLM to .33 (further measures in Supp.~Sec.~\ref{sec:supp_vision}). The frontier MLLMs struggle most with cerebellar lesions and lesions with little oedema, finding at most .377 and .275 of them at 1 FP per scan, compared with .959 and .882 for r15 (Supp.~Tab.~\ref{tab:vision_c}); \cref{fig:hard} shows one of each.

\subsection{What Did \method{} Learn?}
\label{sec:analysis}

\paragraph{Gains accumulate over rounds, and every accepted step holds on held-out data.} Over the 15 rounds, the examiner accepted 8 revisions and rejected 7, and validation $M$ rose from .664/.694/.667 (r0) to .889/.878/.865 (r15) on the three seeds. No single round dominates: the largest step is +.050 (r3), and the four levels take turns (\cref{tab:ablation}). When every accepted system is scored on held-out data after the freeze, each accepted revision also raises test $M$, from .747 through .779, .802, .842, .869, .876, .906 and .920 to .921 (\cref{fig:traj}).

\begin{table}[t]
\centering\small
\caption{\textbf{Accepted revisions and leave-one-out ablation} ($M$, mean of 3 seeds). Acc.: validation $\Delta M$ at acceptance. Each row removes one revision from r15. $^\dagger$Under greedy SFT reads (768\,px: .798/.867/.795).}
\label{tab:ablation}
\setlength{\tabcolsep}{3.5pt}
\begin{tabular}{@{}llcccc@{}}
\toprule
 & Removed revision & Acc. & Val. & Test & Ext. \\
\midrule
-- & none (r15) & & .877 & .921 & .865 \\
r1 EA & 2\,mm $\rightarrow$ 5\,mm survey & +.028 & .808 & .868 & .781 \\
r3 VR & 768 $\rightarrow$ 384\,px$^\dagger$ & +.050 & .758 & .826 & .756 \\
r8 LP & GRPO $\rightarrow$ SFT only & +.018 & .863 & .905 & .834 \\
r12 DA & merge 6 $\rightarrow$ 12\,mm & +.031 & .844 & .891 & .828 \\
r14 DA & blobness $\rightarrow$ peak (r2) & +.015 & .863 & .906 & .836 \\
r5 DA & sampled reads & +.044 & .871 & .918 & .862 \\
r15 DA & one evidence scale & +.002 & .876 & .920 & .864 \\
\bottomrule
\end{tabular}
\end{table}

\paragraph{Every gain comes with its explanation.} Round 12 shows the chain the ledger records for each revision. \emph{Miss:} the census of r8 put the largest share of misses under \emph{not recognized}. \emph{Diagnosis:} the mentor wrote a new diagnostic over the raw points and found that 19 of the 50 lesions missed at 2 FP per scan had a greedy-read point within 5\,mm: \emph{``reports of a small lesion chain into a neighbouring cluster \ldots{} the cluster's medoid then lies $>$5\,mm from the lesion.''} \emph{Hypothesis,} written before the run: a 6\,mm merge radius, \emph{``chosen a priori, no sweep,''} keeps clusters compact relative to the 5\,mm hit tolerance. \emph{Verdict:} $\Delta M{=}{+}.031$ on the deciding seed, positive on all three, accepted; removing it later costs $-.030$ on test (\cref{tab:ablation}).

\paragraph{The misses that drove the loop are the ones it removed.} On the held-out test set at 1 FP per scan, misses fall from 100.3 for r0 to 25.0 for r15 (mean of 3 seeds; Supp.~Fig.~\ref{fig:census}). \emph{Seen, not recognized}, the dominant cause at r0 and the one the mentor targeted most, falls from 82.0 to 18.0, \emph{inconsistent} falls from 18.3 to 4.7, and \emph{not seen} is 0 for both systems. \emph{Rejected} rises from 0 to 2.3, since r15 proposes enough candidates that a few true lesions fall below the budget. BrainMetShare shows the same shift (112.7 to 36.0 misses; \emph{not recognized} 91.0 to 21.0).

\paragraph{Every level contributes.} Removing any of the seven accepted revisions still in r15 lowers $M$ on validation, test and the external cohort (\cref{tab:ablation}; r14 replaced r2's tie-break, so its row reverts to r2). Thin slices matter most on held-out data ($-.053$ on test and $-.084$ on the external cohort).

\paragraph{The frozen rule keeps the loop honest.} The examiner rejected, among others, 1152\,px views, longer GRPO and a 2.5D RGB input, each extending a change that had just helped. The validation gain of .20 carries over to test (.17) and the external cohort (.22). The run took 35\,h and 72.8 GPU-hours (Supp.~Tabs.~\ref{tab:rounds}, \ref{tab:cost}).
\section{Conclusion}
\label{sec:conclusion}
% Answer the central question in 3--4 sentences: yes --- sorted by cause, a system's own misses can drive 15 rounds of
% unattended, audited recursive self-improvement that turn a 4B MLLM into a small-metastasis finder beyond frontier MLLMs,
% robust across institutions and lesion types; the gains accumulate across all four levels, and the changes it keeps are
% the ones radiologists would make.
We presented \method{}, which improves an MLLM at finding small lesions from its own misses: a failure census attributes every miss to a cause, an LLM mentor revises one level per round, and a frozen, audited rule decides what is kept. Over 15 unattended rounds, every accepted revision also raised held-out test $M$, and a 4B MLLM rose from $M = .747$ to $.921$, finding 91\% of lesions of 0.1--1\,mL at 1 FP per scan and surpassing frontier MLLMs on held-out data, an external cohort and an unseen tumour type. Future work will extend the loop to other lesions and imaging modalities and let it revise the vision encoder.
\section*{LLM Usage Statement}
\label{sec:llm_usage}
LLMs are part of the method and of the evaluation. Claude Opus~5.5 acts as the mentor (\cref{sec:formulation}); it proposed, implemented and ran every revision under the frozen protocol, and all of its tool calls were audited (\cref{sec:examiner}). The trainee is Qwen3-VL-4B, and the baselines are the MLLMs listed in \cref{sec:setup}, accessed through their public weights or APIs. Beyond this, Claude was used to edit and polish author-produced text and to write and debug code; the authors checked all text, code and results and take full responsibility for them.

{
    \small
    \bibliographystyle{ieeenat_fullname}
    \bibliography{main}
}

% Works cited only in the supplement are listed at its end as [S1]--[S3], so the main reference list holds only works cited in the main paper.
\newcommand{\suppcite}[1]{[\hyperlink{supref:#1}{#1}]}
\clearpage
\setcounter{page}{1}
\maketitlesupplementary
\renewcommand{\thetable}{S\arabic{table}}
\renewcommand{\thefigure}{S\arabic{figure}}
\setcounter{table}{0}\setcounter{figure}{0}\setcounter{section}{0}
\renewcommand{\thesection}{S\arabic{section}}

% Planned supplementary content (not yet written):
% \section*{Supplementary outline}
% \begin{itemize}
%   \item Mentor brief and tool list; full round ledger (hypothesis, command, result) for r1--r15.
%   \item Frozen acceptance rule and FREEZE hashes; list of logged infrastructure fixes.
%   \item Prompt, rendering and parsing details; baseline settings (models, tokens, sampling).
%   \item Overlap check BraTS-MET vs BrainMetShare; MEN-RT skull-stripping statistics.
%   \item Per-seed tables and bootstrap details; qualitative examples (hits, misses, FPs).
% \end{itemize}

\section*{Contents}
\newcommand{\suppentry}[2]{\noindent\hyperref[#1]{\ref*{#1}\quad #2}\dotfill\pageref*{#1}\par\smallskip}
\newcommand{\supptab}[2]{\noindent\hspace{1.5em}\hyperref[#1]{Table~\ref*{#1}\quad #2}\dotfill\pageref*{#1}\par\smallskip}
\suppentry{sec:supp_impl}{Implementation details}
\supptab{tab:levels}{The four levels and the failure causes they target}
\suppentry{sec:supp_radiology}{Relation to radiological practice}
\suppentry{sec:supp_audit}{Audit and cost}
\suppentry{sec:supp_rounds}{Round ledger}
\supptab{tab:rounds}{All 15 rounds}
\suppentry{sec:supp_cost}{Cost of the run}
\supptab{tab:cost}{Cost of the 15 rounds}
\suppentry{sec:supp_size}{Recall by lesion size}
\supptab{tab:size}{Recall by lesion size at 1 FP per scan}
\suppentry{sec:supp_froc}{Operating curves}
\suppentry{sec:supp_seeds}{Per-seed results and paired bootstrap}
\supptab{tab:seeds}{Per-seed $M$ and paired bootstrap}
\suppentry{sec:supp_infer}{Inference cost}
\supptab{tab:infer}{Cost of reading one test scan}
\suppentry{sec:supp_qual}{Qualitative examples}
\suppentry{sec:supp_vision}{Finer visual measures}
\supptab{tab:vision_a}{Slice-level detection and localisation}
\supptab{tab:vision_b}{Recall by scan and lesion properties}
\suppentry{sec:supp_census}{Failure census on held-out data}
\suppentry{sec:supp_so}{Score-only ablation}
\clearpage

\section{Implementation details}
\label{sec:supp_impl}
\Cref{tab:levels} lists the four revision levels of the main paper and the failure causes each one targets.

\begin{table}[h]
\centering
\small
\setlength{\tabcolsep}{4pt}
\begin{tabular}{@{}>{\raggedright\arraybackslash\hyphenpenalty=10000}p{0.25\linewidth}>{\raggedright\arraybackslash\hyphenpenalty=10000}p{0.36\linewidth}>{\raggedright\arraybackslash\hyphenpenalty=10000}p{0.31\linewidth}@{}}
\toprule
Level & What it changes & Targets \\
\midrule
Evidence Acquisition & views, number of reads & not seen, not recognized \\ \addlinespace[3pt]
Visual Representation & render size, input encoding & not recognized \\ \addlinespace[3pt]
Learning Policy & training data, objective & not recognized, inconsistent \\ \addlinespace[3pt]
Decision Aggregation & merging, ranking & inconsistent, rejected \\
\bottomrule
\end{tabular}
\caption{\textbf{The four levels} and the failure causes each one targets.}
\label{tab:levels}
\end{table}

\paragraph{SFT (r0).} LoRA adapters~\cite{hu2022lora} (rank 16, $\alpha{=}32$, dropout 0.05) on all attention and MLP projections of the language model, vision encoder frozen; 2 epochs, learning rate $2{\times}10^{-4}$, cosine schedule with 3\% warm-up, batch size 4.
\paragraph{GRPO (r8).} 600 steps; each step draws 16 training views (half with a lesion) and samples $G{=}4$ reports per view at temperature 1. Unparsable output receives $R_i{=}{-}1$. The loss is the advantage-weighted mean token log-likelihood, without a KL term, at learning rate $10^{-5}$.
\paragraph{Blobness (r14).} At scale 2\,mm, $\beta$ is non-zero only where all three Hessian eigenvalues are negative, and is multiplied by the ratio of the smallest to the largest eigenvalue magnitude.
\paragraph{Census diagnostics.} Besides the four counts, the census reports the number of surveyed slices that intersect each missed lesion and the support of rejected hits compared with that of FPs.

\section{Relation to radiological practice}
\label{sec:supp_radiology}
Without being told, the mentor moved to thin slices (5$\rightarrow$2\,mm), as consensus protocols for brain metastases prescribe thin-section 3D post-contrast T1-weighted imaging~\suppcite{S1} and thin-slice post-contrast imaging detects more small metastases~\suppcite{S2}; magnified the view (384$\rightarrow$768\,px); and added a compact-lesion prior (Hessian blobness), which favours nodular enhancing foci, the typical appearance of metastases~\suppcite{S2}, over tubular structures such as vessels~\cite{frangi1998multiscale}, with a merge radius matching small-lesion size.

\section{Audit and cost}
\label{sec:supp_audit}
373 tool calls; no access to test/external data, credentials or network; scoring functions unchanged (AST-identical); every infrastructure fix is recorded in a separate log. The 15 rounds took 35\,h of wall-clock time and 72.8 GPU-hours on single GPUs; 7 of 15 rounds needed no GPU because they re-aggregated saved reads (\cref{tab:cost}). The mentor made 217 model calls with 0.2M output tokens. The final trainee is a 4B model that reads locally; frontier MLLM baselines are billed per scan at inference.

\section{Round ledger}
\label{sec:supp_rounds}
\Cref{tab:rounds} lists every round, accepted or rejected, with the validation $\Delta M$ on which the frozen rule decided.

\begin{table}[t]
\centering\small
\caption{\textbf{All 15 rounds.} $\Delta M$ on validation as used by the frozen rule (mean of 3 seeds for revisions that re-read or retrain; first seed for aggregation on saved reads).}
\label{tab:rounds}
\setlength{\tabcolsep}{4pt}
\begin{tabular}{@{}llr>{\raggedright\arraybackslash}p{0.52\linewidth}@{}}
\toprule
Round & Level & $\Delta M$ & Revision (decision) \\
\midrule
r1 & EA & +.0284 & survey spacing 5$\rightarrow$2\,mm (accept) \\
r2 & DA & +.0234 & peak-intensity tie-break (accept) \\
r3 & VR & +.0495 & 384$\rightarrow$768\,px (accept) \\
r4 & EA & $-$.0023 & sampled reads, pairwise ranking (reject) \\
r5 & DA & +.0438 & greedy-first + 4 sampled reads (accept) \\
r6 & DA & $-$.0006 & sampled-confirmation tie-break (reject) \\
r7 & VR & $-$.0118 & 1152\,px (reject) \\
r8 & LP & +.0179 & GRPO on top of SFT (accept) \\
r9 & LP & $-$.0015 & GRPO continuation, $G{=}8$ (reject) \\
r10 & DA & $-$.0006 & two-model report (reject) \\
r11 & VR & $-$.0164 & 2.5D RGB input (reject) \\
r12 & DA & +.0310 & merge radius 12$\rightarrow$6\,mm (accept) \\
r13 & EA & $-$.0015 & 1\,mm greedy read (reject) \\
r14 & DA & +.0145 & 3D Hessian blobness (accept) \\
r15 & DA & +.0019 & one evidence scale across tiers (accept) \\
\bottomrule
\end{tabular}
\end{table}

\section{Cost of the run}
\label{sec:supp_cost}
\Cref{tab:cost} breaks down the 72.8 GPU-hours of the run by round; the seven aggregation rounds ran on CPU only.

\begin{table}[t]
\centering\small
\caption{\textbf{Cost of the 15 rounds.} GPU-hours summed over the 3 seeds; $^*$r2, r5, r6, r10, r12, r14, r15, run on CPU.}
\label{tab:cost}
\setlength{\tabcolsep}{4pt}
\begin{tabular}{@{}llr@{}}
\toprule
Rounds & Work & GPU-h \\
\midrule
r1 / r13 & re-read at 2 / 1\,mm & 0.6 / 1.1 \\
r3 / r7 & SFT + read, 768 / 1152\,px & 9.2 / 22.2 \\
r4 & 4 sampled reads & 2.1 \\
r8 / r9 & GRPO / continuation & 9.4 / 10.7 \\
r11 & SFT + read, 2.5D RGB & 17.5 \\
7 DA rounds$^*$ & aggregation of saved reads & 0 \\
\midrule
Total & 15 rounds, 35\,h & 72.8 \\
\bottomrule
\end{tabular}
\end{table}

\section{Recall by lesion size}
\label{sec:supp_size}
\Cref{tab:size} gives recall at 1 FP per scan for every lesion-size range, baseline and dataset.

\begin{table*}[t]
\centering\small
\caption{\textbf{Recall by lesion size} at 1 FP per scan (lesion volume in mL). \method{}: mean of 3 seeds. The gap between r15 and every baseline is largest on the smallest lesions.}
\label{tab:size}
\setlength{\tabcolsep}{3.5pt}
\begin{tabular}{l cccc cccc cccc}
\toprule
& \multicolumn{4}{c}{BraTS-MET test} & \multicolumn{4}{c}{BrainMetShare (external)} & \multicolumn{4}{c}{MEN-RT (unseen type)} \\
\cmidrule(lr){2-5}\cmidrule(lr){6-9}\cmidrule(lr){10-13}
Model & .05--.1 & .1--1 & 1--5 & $>$5 & .05--.1 & .1--1 & 1--5 & $>$5 & .05--.1 & .1--1 & 1--5 & $>$5 \\
\midrule
MedGemma 27B & .000 & .000 & .000 & .000 & .000 & .000 & .021 & .000 & .000 & .008 & .005 & .014 \\
Qwen3-VL-4B zero-shot & .000 & .004 & .122 & .648 & .005 & .016 & .146 & .462 & .000 & .000 & .025 & .469 \\
Gemma~4 31B & .000 & .104 & .610 & .796 & .005 & .103 & .604 & .692 & .000 & .040 & .323 & .767 \\
Qwen3.8-27B & .000 & .108 & .585 & .963 & .005 & .099 & .646 & .846 & .000 & .016 & .283 & .760 \\
GPT-6.1 Sol & .009 & .065 & .610 & .981 & .000 & .063 & .625 & .923 & .000 & .040 & .237 & .757 \\
Claude Opus~5.5 & .017 & .292 & .902 & .981 & .016 & .273 & .938 & .846 & .000 & .024 & .258 & .805 \\
Gemini~3.7 Flash & .017 & .285 & .915 & .981 & .021 & .265 & .958 & .923 & .000 & .160 & .465 & .839 \\
\midrule
\method{} r0 & .151 & .629 & .951 & 1.00 & .147 & .569 & .924 & 1.00 & .037 & .280 & .576 & .849 \\
\method{} r15 & \textbf{.638} & \textbf{.908} & \textbf{.988} & \textbf{1.00} & \textbf{.558} & \textbf{.862} & \textbf{.979} & \textbf{1.00} & \textbf{.130} & \textbf{.573} & \textbf{.695} & \textbf{.903} \\
\bottomrule
\end{tabular}
\end{table*}

\section{Operating curves}
\label{sec:supp_froc}
\begin{figure}[t]
\centering
\includegraphics[width=\linewidth]{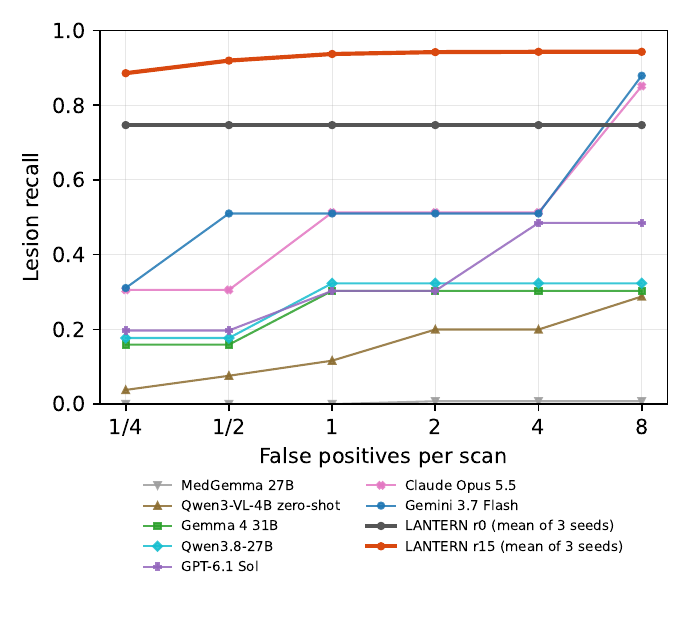}
\caption{\textbf{Operating curves on the held-out test set} (lesions $\geq$0.1\,mL). \method{}: mean of 3 seeds.}
\label{fig:supp_froc}
\end{figure}

\Cref{fig:supp_froc} shows lesion recall on the held-out test set at 0.25 to 8 false positives per scan. $M$ averages the first four budgets. r15 leads at every budget; the two strongest frontier MLLMs exceed r0 only at 8 false positives per scan.

\section{Per-seed results and paired bootstrap}
\label{sec:supp_seeds}
\Cref{tab:seeds} lists $M$ of r0 and r15 for each training seed. For held-out data, $\Delta M$ is compared on the same seed with a paired patient-level bootstrap (2{,}000 resamples), the same procedure the examiner logs during the run; P is the fraction of resamples with $\Delta M>0$. During the run, the examiner also logs for every revision whether all three seeds improved and this bootstrap probability of improvement; neither enters the acceptance rule (\cref{eq:rule}).

\begin{table}[t]
\centering\small
\caption{\textbf{Per-seed $M$ of r0 and r15} and the paired patient-level bootstrap of $\Delta M$ (95\% interval).}
\label{tab:seeds}
\setlength{\tabcolsep}{3.5pt}
\begin{tabular}{@{}llccccc@{}}
\toprule
Data & Seed & r0 & r15 & $\Delta M$ & 95\% interval & P \\
\midrule
Validation & 20 & .664 & .889 & +.225 & -- & -- \\
 & 21 & .694 & .878 & +.184 & -- & -- \\
 & 22 & .667 & .865 & +.198 & -- & -- \\
\midrule
Test & 20 & .740 & .929 & +.189 & [.126, .249] & 1.00 \\
 & 21 & .768 & .920 & +.152 & [.091, .230] & 1.00 \\
 & 22 & .732 & .914 & +.181 & [.116, .257] & 1.00 \\
\midrule
BrainMetShare & 20 & .624 & .870 & +.246 & [.190, .309] & 1.00 \\
 & 21 & .669 & .862 & +.193 & [.150, .316] & 1.00 \\
 & 22 & .631 & .862 & +.232 & [.178, .291] & 1.00 \\
\midrule
MEN-RT & 20 & .605 & .750 & +.146 & [.087, .171] & 1.00 \\
 & 21 & .619 & .739 & +.120 & [.090, .155] & 1.00 \\
 & 22 & .598 & .732 & +.134 & [.090, .159] & 1.00 \\
\bottomrule
\end{tabular}
\end{table}

\section{Inference cost}
\label{sec:supp_infer}
\Cref{tab:infer} compares the cost of reading one test scan. r15 reads more views than the baselines but runs locally; the API baselines are billed per token at list prices (November 2026).

\begin{table}[t]
\centering\small
\caption{\textbf{Cost of reading one test scan.} Views at the system's own protocol; tokens are measured averages per view; API cost at standard list prices. r15 (4B) runs locally on half an H100 NVL.}
\label{tab:infer}
\setlength{\tabcolsep}{3pt}
\begin{tabular}{@{}lccl@{}}
\toprule
System & Views $\times$ reads & Tokens in / out & Cost per scan \\
\midrule
GPT-6.1 Sol & 29 $\times$ 1 & 261 / 164 & \$0.06 \\
Gemini~3.7 Flash & 29 $\times$ 1 & 1{,}218 / 495 & \$0.08 \\
Claude Opus~5.5 & 29 $\times$ 1 & 358 / 190 & \$0.15 \\
\method{} r15 & 76 $\times$ 5 & -- & local, 25\,s \\
\bottomrule
\end{tabular}
\end{table}

\section{Qualitative examples}
\label{sec:supp_qual}
\begin{figure}[t]
\centering
\includegraphics[width=0.9\linewidth]{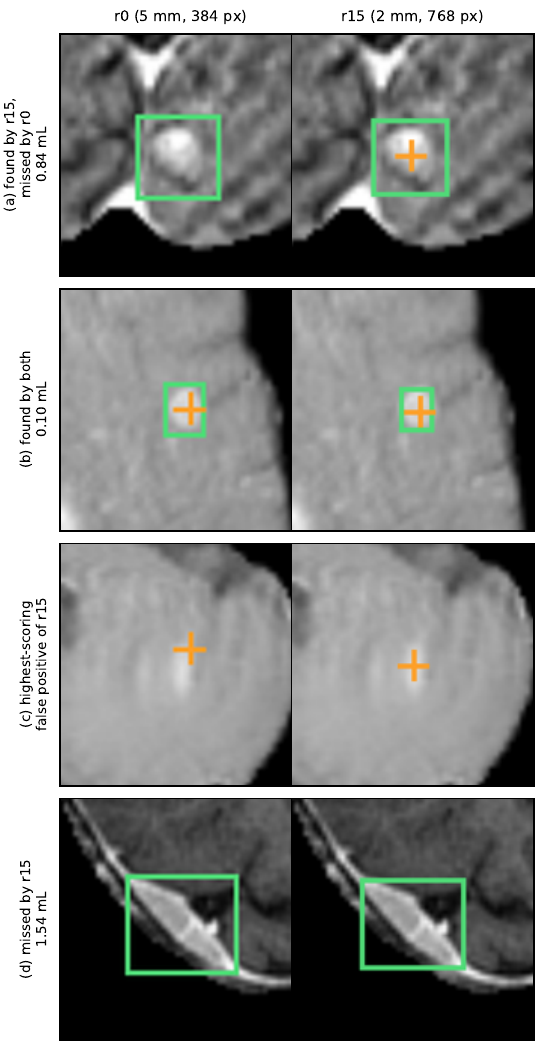}
\caption{\textbf{r0 vs.\ r15 on held-out test scans} (seed 20). Green box: ground-truth lesion on the slice; orange $+$: candidate within the 1 FP per scan budget; grey $+$: candidate outside it.}
\label{fig:supp_qual}
\end{figure}

\Cref{fig:supp_qual} shows four test cases read by r0 and r15 (seed 20), chosen by fixed rules: (a) the largest lesion of 0.1--1\,mL that r15 finds and r0 misses, (b) the smallest lesion both find, (c) the highest-scoring false positive of r15 within the budget, and (d) the largest lesion r15 misses.

\section{Finer visual measures}
\label{sec:supp_vision}
\Cref{tab:vision_a,tab:vision_b} compare r15 with the frontier MLLMs (and, in \cref{tab:vision_b}, with r0) on the held-out test set, from the saved reads (no re-reading). \emph{Slice level}: a read view is positive when a lesion $\geq$0.1\,mL crosses it (views crossed only by smaller lesions are excluded), and a reader calls it positive when it reports any point; specificity and the Youden index are computed over each system's own views (greedy read for r15). \emph{Localisation}: median 3D distance from each hit within the 1 FP per scan budget to the centroid of the lesion it hits. \Cref{tab:vision_b} gives recall at 1 FP per scan by the number of lesions in the scan, by location (lesion centroid within 10\,mm of the brain surface or deeper; inferior third of the brain or the rest) and by contrast (lesion mean T1c minus the median of a 2--5\,mm shell, in quartiles). \method{}: mean of 3 seeds.

\begin{table}[t]
\centering\small
\caption{\textbf{Finer visual measures, all baselines}, test set (\cref{tab:vision} of the main paper with the open-weight MLLMs). J: slice-level Youden index. Dist.: median hit-to-centroid distance (mm). D50: diameter (mm) found with 50\% probability. Recall at 1 FP per scan in scans with $\geq$10 lesions, near the surface (Surf., $\leq$10\,mm), deeper (Deep) and in the lowest-contrast quartile (Low c.). --: undefined (MedGemma finds no lesion at 1 FP per scan; Gemma~4 recall stays below 50\% at every size).}
\label{tab:vision_full}
\setlength{\tabcolsep}{1.8pt}
\resizebox{\linewidth}{!}{%
\begin{tabular}{@{}lcccccccc@{}}
\toprule
 & J$\uparrow$ & \multicolumn{2}{c}{Dist.$\downarrow$} & D50$\downarrow$ & $\geq$10$\uparrow$ & Surf.$\uparrow$ & Deep$\uparrow$ & Low c.$\uparrow$ \\
\cmidrule(lr){3-4}
Model & & all & $<$1\,mL & & & & & \\
\midrule
MedGemma 27B & .137 & -- & -- & -- & .000 & .000 & .000 & .000 \\
Gemma~4 31B & .598 & 4.68 & 4.50 & -- & .163 & .200 & .359 & .303 \\
Qwen3.8-27B & .554 & 7.11 & 5.79 & 12.2 & .200 & .193 & .395 & .343 \\
GPT-6.1 Sol & .239 & \underline{1.92} & \underline{1.54} & 7.7 & .212 & .221 & .348 & .293 \\
Claude Opus~5.5 & .552 & 2.69 & 2.88 & 9.7 & .325 & .393 & .578 & .475 \\
Gemini~3.7 Flash & .687 & 2.88 & 2.99 & 9.7 & .300 & .443 & .547 & .465 \\
\midrule
\method{} r0 & \underline{.706} & 2.02 & 1.74 & \underline{7.1} & \underline{.633} & \underline{.624} & \underline{.814} & \underline{.609} \\
\method{} r15 & \textbf{.798} & \textbf{1.22} & \textbf{0.99} & \textbf{4.9} & \textbf{.908} & \textbf{.893} & \textbf{.961} & \textbf{.875} \\
\bottomrule
\end{tabular}}
\end{table}

\begin{table}[t]
\centering\small
\caption{\textbf{Slice-level detection and localisation} on the test set. Spec.: slice-level specificity; $J$: Youden index (sensitivity\,+\,specificity\,$-$\,1); Dist.: median distance (mm) from a hit to the lesion centroid, all lesions and lesions $<$1\,mL.}
\label{tab:vision_a}
\setlength{\tabcolsep}{3pt}
\begin{tabular}{@{}lcccc@{}}
\toprule
 & Spec. & $J$ & Dist. & Dist. $<$1\,mL \\
\midrule
\method{} r15 & \textbf{.983} & \textbf{.798} & \textbf{1.22} & \textbf{0.99} \\
Gemini~3.7 Flash & .813 & .687 & 2.88 & 2.99 \\
Claude Opus~5.5 & .648 & .552 & 2.69 & 2.87 \\
GPT-6.1 Sol & .293 & .239 & 1.92 & 1.54 \\
Qwen3.8-27B & .758 & .554 & 7.11 & 5.79 \\
Gemma~4 31B & .813 & .598 & 4.68 & 4.50 \\
MedGemma 27B & .213 & .137 & -- & -- \\
Qwen3-VL-4B zero-shot & .253 & .197 & 9.18 & 6.56 \\
\bottomrule
\end{tabular}
\end{table}

\begin{table*}[t]
\centering\small
\caption{\textbf{Recall at 1 FP per scan by scan and lesion properties} on the test set (lesions $\geq$0.1\,mL).}
\label{tab:vision_b}
\setlength{\tabcolsep}{3.5pt}
\begin{tabular}{@{}l cccc cc cc cccc@{}}
\toprule
 & \multicolumn{4}{c}{Lesions per scan} & \multicolumn{2}{c}{Depth} & \multicolumn{2}{c}{Height} & \multicolumn{4}{c}{Contrast quartile} \\
\cmidrule(lr){2-5}\cmidrule(lr){6-7}\cmidrule(lr){8-9}\cmidrule(lr){10-13}
 & 1 & 2--3 & 4--9 & $\geq$10 & $\leq$10\,mm & deeper & inferior & rest & Q1 & Q2 & Q3 & Q4 \\
\midrule
\method{} r0 & .882 & .785 & .744 & .633 & .624 & .814 & .725 & .757 & .609 & .747 & .801 & .828 \\
\method{} r15 & \textbf{.938} & \textbf{.945} & \textbf{.945} & \textbf{.908} & \textbf{.893} & \textbf{.961} & \textbf{.929} & \textbf{.941} & \textbf{.875} & \textbf{.960} & \textbf{.960} & \textbf{.953} \\
Gemini~3.7 Flash & .688 & .620 & .508 & .300 & .443 & .547 & .389 & .570 & .465 & .364 & .515 & .697 \\
Claude Opus~5.5 & .667 & .582 & .524 & .325 & .393 & .578 & .374 & .581 & .475 & .374 & .525 & .677 \\
GPT-6.1 Sol & .542 & .367 & .254 & .212 & .221 & .348 & .176 & .366 & .293 & .152 & .283 & .485 \\
Qwen3.8-27B & .479 & .380 & .312 & .200 & .193 & .395 & .214 & .377 & .343 & .182 & .293 & .475 \\
Gemma~4 31B & .500 & .316 & .307 & .163 & .200 & .359 & .198 & .355 & .303 & .172 & .283 & .455 \\
MedGemma 27B & .000 & .000 & .000 & .000 & .000 & .000 & .000 & .000 & .000 & .000 & .000 & .000 \\
Qwen3-VL-4B zero-shot & .354 & .165 & .074 & .025 & .029 & .164 & .015 & .166 & .131 & .040 & .121 & .172 \\
\bottomrule
\end{tabular}
\end{table*}

\Cref{tab:vision_c} gathers the hardest lesion groups on the test set and repeats the size and contrast measures on the two external cohorts. A lesion is \emph{crowded} when the edge-to-edge gap to its nearest other lesion ($\geq$0.1\,mL) is at most 10\,mm; \emph{Pair} is the share of such close pairs with both lesions found. \emph{Solid}: fewer than 5 voxels carry the BraTS necrosis label. \emph{Oed.}: peritumoural oedema within 10\,mm of the lesion is at most 1\,mL. \emph{Cbl.}: cerebellum, from the CerebrA atlas~\suppcite{S3} registered to each scan. D50 is left blank when the fitted value lies beyond the largest lesion of the cohort.

\begin{table*}[t]
\centering\small
\caption{\textbf{Recall on the hardest lesion groups} at 1 FP per scan (lesions $\geq$0.1\,mL; D50 over all lesions). Test: held-out BraTS-MET. BMS: BrainMetShare. MEN: BraTS-MEN-RT meningiomas, read zero-shot. Low c.: lowest-contrast quartile; $\geq$10: scans with ten or more lesions; Surf.: centroid within 10\,mm of the brain surface. Number of test lesions: crowded 58, solid 225, Oed.\ 200, Cbl.\ 106; 38 close pairs. \method{}: mean of 3 seeds.}
\label{tab:vision_c}
\setlength{\tabcolsep}{2.6pt}
\resizebox{\linewidth}{!}{%
\begin{tabular}{@{}l ccccccccc ccc ccc@{}}
\toprule
 & \multicolumn{9}{c}{Test} & \multicolumn{3}{c}{BMS} & \multicolumn{3}{c}{MEN} \\
\cmidrule(lr){2-10}\cmidrule(lr){11-13}\cmidrule(lr){14-16}
 & D50$\downarrow$ & Low c. & Surf. & $\geq$10 & Crowded & Pair & Solid & Oed. & Cbl. & D50$\downarrow$ & Low c. & $\geq$10 & D50$\downarrow$ & Low c. & $\geq$10 \\
\midrule
\method{} r0 & 7.1 & .609 & .624 & .633 & .730 & .614 & .639 & .597 & .745 & 7.3 & .574 & .680 & 16.5 & .481 & .440 \\
\method{} r15 & \textbf{4.9} & \textbf{.875} & \textbf{.893} & \textbf{.908} & \textbf{.937} & \textbf{.912} & \textbf{.913} & \textbf{.882} & \textbf{.959} & \textbf{5.0} & \textbf{.755} & \textbf{.889} & \textbf{9.6} & \textbf{.632} & \textbf{.821} \\
Gemini~3.7 Flash & 9.7 & .465 & .443 & .300 & .483 & .263 & .338 & .275 & .368 & 8.7 & .405 & .347 & 40.2 & .409 & .429 \\
Claude Opus~5.5 & 9.7 & .475 & .393 & .325 & .448 & .263 & .333 & .275 & .377 & 10.4 & .443 & .347 & 38.1 & .338 & .107 \\
GPT-6.1 Sol & 7.7 & .293 & .221 & .212 & .379 & .184 & .147 & .055 & .198 & 14.2 & .177 & .187 & 11.9 & .299 & .107 \\
Qwen3.8-27B & 12.2 & .343 & .193 & .200 & .379 & .289 & .169 & .110 & .255 & 11.7 & .152 & .173 & 11.8 & .305 & .107 \\
Gemma~4 31B & -- & .303 & .200 & .163 & .345 & .237 & .187 & .115 & .217 & -- & .203 & .133 & 40.0 & .338 & .286 \\
MedGemma 27B & -- & .000 & .000 & .000 & .000 & .000 & .000 & .000 & .000 & -- & .013 & .000 & -- & .006 & .036 \\
Qwen3-VL-4B zero-shot & -- & .131 & .029 & .025 & .086 & .000 & .044 & .000 & .009 & 25.1 & .025 & .013 & -- & .123 & .036 \\
\bottomrule
\end{tabular}}
\end{table*}

\begin{figure*}[t]
\centering
\includegraphics[width=\linewidth]{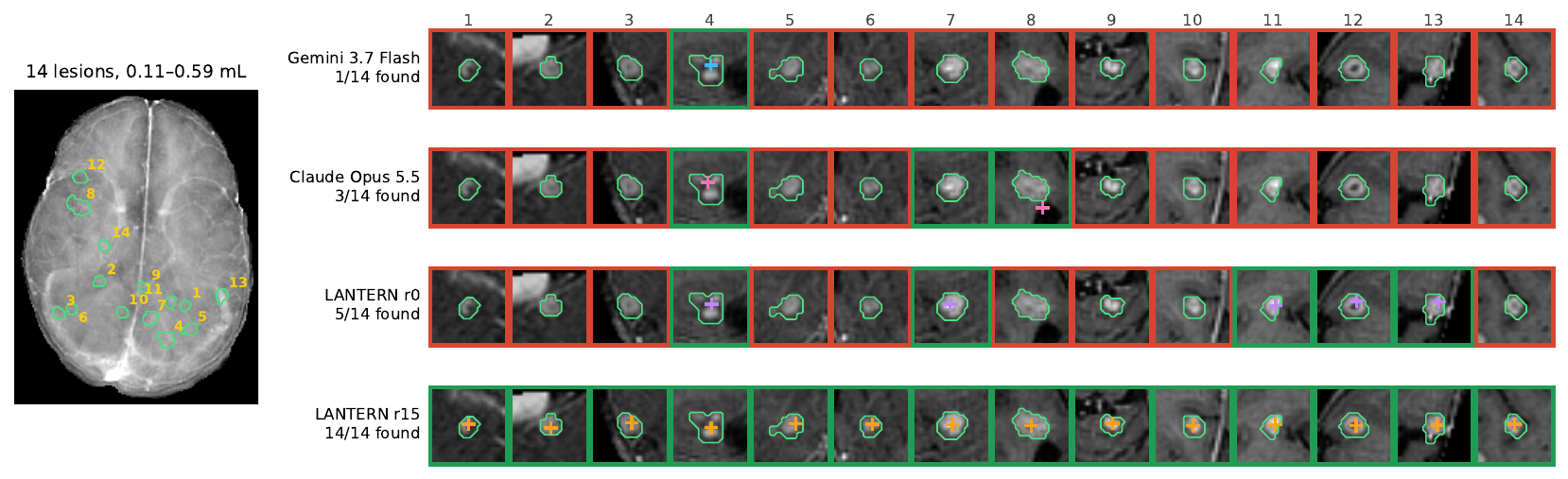}
\caption{\textbf{Fourteen small metastases in one held-out test scan} (T1c). Left: mean-intensity projection, every lesion $\geq$0.1\,mL outlined and numbered from inferior to superior. Right: each lesion on its centroid slice, one row per system (seed 20 for \method{}); frame green, found; red, missed at 1 FP per scan; $+$, candidates the system keeps within two slices. A hit is any point within 5\,mm of the lesion mask in 3D, so a hit can lie outside the outline drawn on one slice (Claude Opus~5.5, lesion 8).}
\label{fig:many_mets}
\end{figure*}

Beyond the two examples of \cref{fig:hard}, \cref{fig:many_mets} shows a whole scan: the test scan with ten or more lesions that r15 reads most completely.

\section{Failure census on held-out data}
\label{sec:supp_census}
\begin{figure*}[h]
\centering
\includegraphics[width=\textwidth]{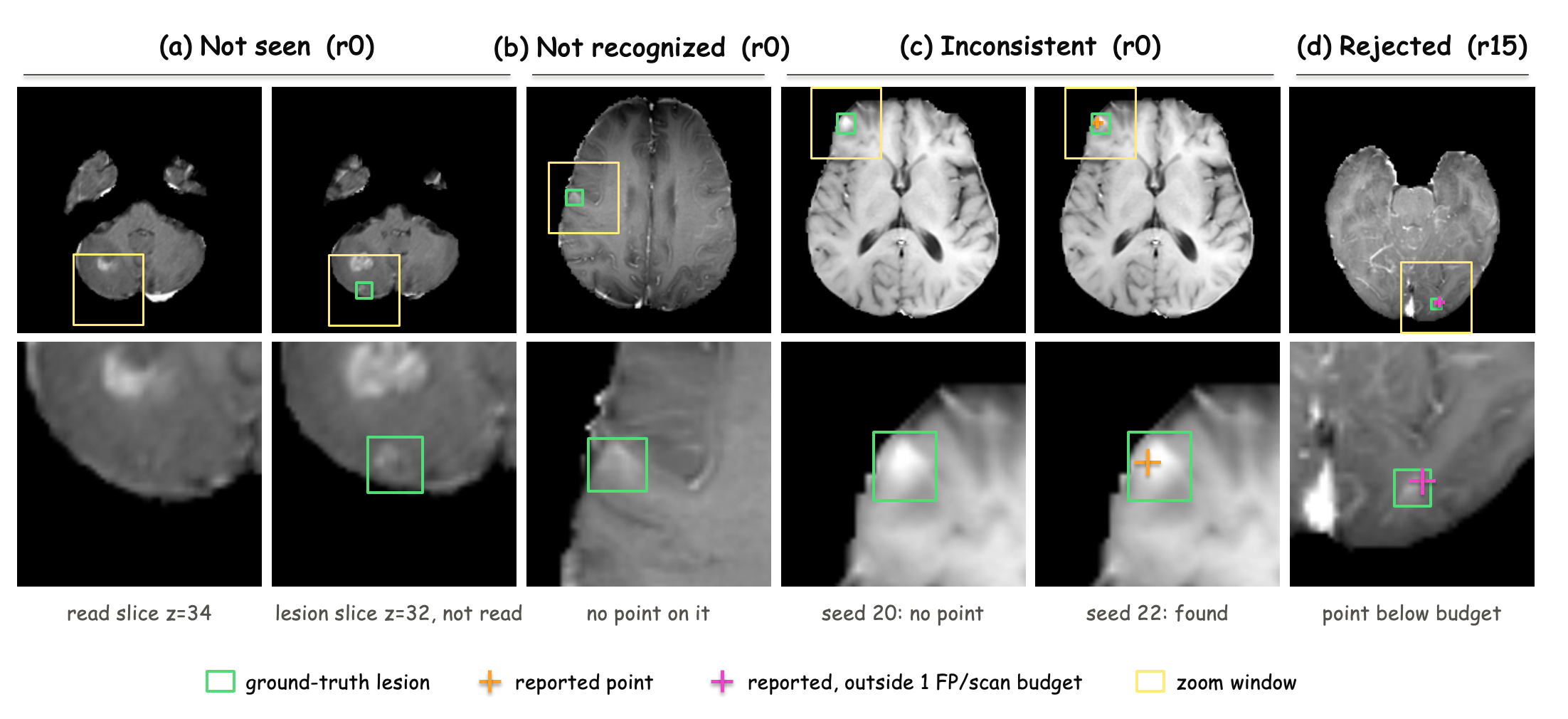}
\caption{\textbf{The four failure causes on real validation scans.} Top: the full slice the system read (yellow: zoom window). Bottom: zoom on the lesion. Green box: ground-truth lesion on that slice; orange $+$: reported point; magenta $+$: reported point whose candidate falls outside the 1 FP per scan budget. (a) The 5\,mm survey reads $z{=}34$; the lesion spans $z{=}30$--$33$. (b) The lesion is on a read slice, and no seed reports it. (c) Seed~20 misses a lesion that seed~22 finds on the same slice. (d) r15 reports the lesion, but its candidate ranks below the budget. (a)--(c) are from r0 and (d) from r15, since r0 never fills its FP budget.}
\label{fig:failure}
\end{figure*}

\Cref{fig:failure} shows one real validation example of each failure cause.

\begin{figure}[h]
\centering
\includegraphics[width=0.85\linewidth]{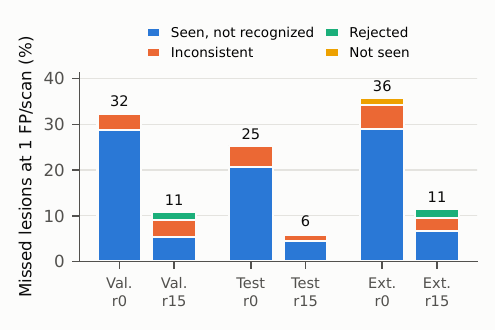}
\caption{\textbf{Failure census, r0 vs.\ r15}: missed lesions $\geq$0.1\,mL at 1 FP per scan (\% of lesions, mean of 3 seeds).}
\label{fig:census}
\end{figure}

\Cref{fig:census} applies the failure census of the main paper to the validation set, the held-out test set and the external cohort, for r0 and r15. The census of the test and external data was computed once, after the run was frozen, and never informed a decision.

\section{Score-only ablation}
\label{sec:supp_so}
To test whether the error diagnostics are needed, we reran the loop with a mentor that sees only the score. Everything else was unchanged: the same r0, trainee, mentor (Claude Opus~5.5), examiner, levels, acceptance and stop rules, and 15-round budget. The validation labels were removed from the mentor's workspace; a scorer outside it returned only $M$ and recall and FP per scan at the four budgets of $M$, and the failure census was not available. Starting from the same r0, \method{} reached the validation $M$ of the final score-only system in 8 rounds instead of 15.

\section*{Supplementary References}
{\small
\noindent Bracketed numbers without ``S'' refer to the reference list of the main paper.

\noindent\hypertarget{supref:S1}{[S1]} Timothy~J. Kaufmann, Marion Smits, Jerrold Boxerman, Raymond Huang, Daniel~P. Barboriak, Michael Weller, et~al. Consensus recommendations for a standardized brain tumor imaging protocol for clinical trials in brain metastases. \emph{Neuro-Oncology}, 22(6):757--772, 2020.

\noindent\hypertarget{supref:S2}{[S2]} Kathleen~R. Fink and James~R. Fink. Imaging of brain metastases. \emph{Surgical Neurology International}, 4(Suppl 4):S209--S219, 2013.

\noindent\hypertarget{supref:S3}{[S3]} Ana~L. Manera, Mahsa Dadar, Vladimir Fonov, and D.~Louis Collins. {CerebrA}, registration and manual label correction of {Mindboggle-101} atlas for {MNI-ICBM152} template. \emph{Scientific Data}, 7(1):237, 2020.
}

\end{document}